\documentclass{article} 
\usepackage[preprint]{colm2025_conference}

\usepackage{microtype}
\usepackage{hyperref}
\usepackage{url}
\usepackage{booktabs}
\usepackage{amsmath}
\usepackage{amssymb}
\usepackage{tabularx}

\usepackage{colortbl}    

\usepackage{times}
\usepackage{float}
\usepackage{latexsym}
\usepackage{booktabs}
\usepackage{amsmath}
\usepackage{hyperref}    
\usepackage{amssymb}     
\usepackage{tipa}
\usepackage{multirow}
\usepackage{xcolor}
\definecolor{mypink}{rgb}{0.9686274,0.882352,0.86666}
\definecolor{mygreen}{rgb}{0.819607,0.890196,0.760784}

\usepackage{latexsym}
\usepackage{graphicx}
\usepackage{float}
\usepackage{colortbl}
\usepackage{multirow}
\usepackage{booktabs}
\usepackage{subcaption}
\usepackage{rotating}
\usepackage{placeins}
\usepackage{comment}
\usepackage{pifont}

\usepackage[utf8]{inputenc}
\DeclareUnicodeCharacter{1E24}{\d{H}}
\DeclareUnicodeCharacter{1E24}{\d{H}}
\usepackage{lineno}

\definecolor{darkblue}{rgb}{0, 0, 0.5}
\hypersetup{colorlinks=true, citecolor=darkblue, linkcolor=darkblue, urlcolor=darkblue}
\usepackage[T1]{fontenc}

\newcommand{\cmark}{\ding{51}} 

\title{Unveiling Factors for Enhanced POS Tagging: \\ A Study of Low-Resource Medieval Romance Languages}

\author{\textbf{Matthias Schöffel}$^{1,2}$ \quad
\textbf{Esteban Garces Arias}$^{2,3}$ \quad
\textbf{Marinus Wiedner}$^{4}$ \quad
\textbf{Paula Ruppert}$^{2}$ \\
\textbf{Meimingwei Li}$^{2}$ \quad
\textbf{Christian Heumann}$^{2}$ \quad
\textbf{Matthias Aßenmacher}$^{2,3}$ \\
\\
$^{1}$Bavarian Academy of Sciences, Munich, Germany \\
$^{2}$LMU Munich, Munich, Germany \\
$^{3}$Munich Center for Machine Learning (MCML), Munich, Germany \\
$^{4}$University of Freiburg, Freiburg, Germany \\
\\
\texttt{matthias.schoeffel@badw.de}
}

\begin{document}

\ifcolmsubmission
\linenumbers
\fi

\maketitle

\begin{abstract}
Part-of-speech (POS) tagging remains a foundational component in natural language processing pipelines, particularly critical for historical text analysis at the intersection of computational linguistics and digital humanities. Despite significant advancements in modern large language models (LLMs) for ancient languages, their application to Medieval Romance languages presents distinctive challenges stemming from diachronic linguistic evolution, spelling variations, and labeled data scarcity. This study systematically investigates the central determinants of POS tagging performance across diverse corpora of Medieval Occitan, Medieval Spanish, and Medieval French texts, spanning biblical, hagiographical, medical, and dietary domains.
Through rigorous experimentation, we evaluate how fine-tuning approaches, prompt engineering, model architectures, decoding strategies, and cross-lingual transfer learning techniques affect tagging accuracy. Our results reveal both notable limitations in LLMs' ability to process historical language variations and non-standardized spelling, as well as promising specialized techniques that effectively address the unique challenges presented by low-resource historical languages.
\end{abstract}

\section{Introduction}
\begin{figure*}[ht]
\centering
\begin{minipage}[c]{0.45\textwidth}
    \centering
    \includegraphics[width=0.95\textwidth]{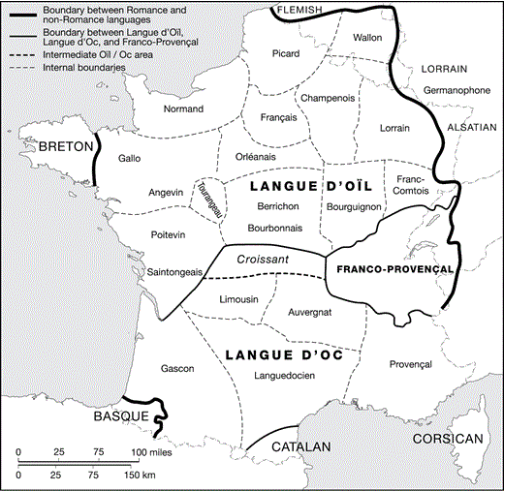}
    \caption*{(a) Overview of the French and Occitan varieties in France \citep{caron:2024}}
    
    \vspace{1.2em}
    
    \includegraphics[width=0.95\textwidth]{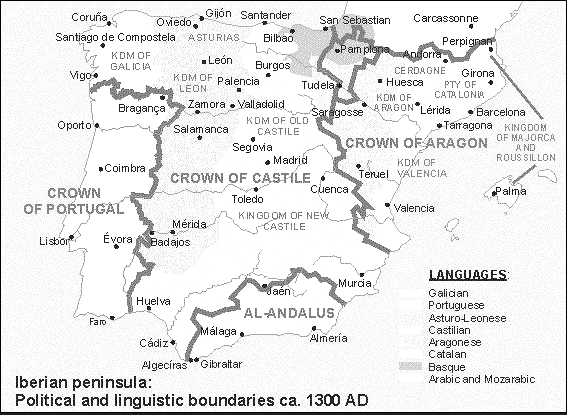}
\caption*{(b) Political linguistic map of Spain around 1300 AD\footnote{Source: "Linguistic and Cultural Diversity of Medieval Iberia (900-1500)", Ballandalus, 2014.}}
\end{minipage}%
\hfill
\begin{minipage}[c]{0.5\textwidth}
    \centering
    \small 
    \begin{tabular}{p{0.35\textwidth}p{0.6\textwidth}}
        \hline
        \textbf{Characteristic} & \textbf{Examples}  \\ 
        \midrule
        \midrule
        Spelling variation & tepms (engl. 'time') vs. tems, temps; tuig (engl. 'all') vs. tug, tot \\ 
        Ambiguous forms & que (CCONJ, SCONJ, PRON) depending on the context\\
        Arabic influences & taxmir (engl. 'blepharoplasty'), written as atactini, ataxmir, tactimi, tactinir, taxani\\
        \bottomrule
    \end{tabular}
    \caption*{(c) Characteristics and examples of linguistic challenges in Medieval Occitan}
    
    \vspace{2.7em}
    
    \small 
    \begin{tabular}{p{0.35\textwidth}p{0.6\textwidth}}
        \hline
        \textbf{Characteristic} & \textbf{Examples}  \\ 
        \midrule
        \midrule
        Middle french traits (1350-1500) & sanc, saunc (engl. 'blood') > sang \\ 
        Spelling variants & muscle, muscule, musculle (engl. 'muscle' ) \\
        Variation in gender assignment & la/ le quantité (engl. 'quantity', f. or m.)\\
        \bottomrule
    \end{tabular}
    \caption*{(d) Characteristics and examples of linguistic challenges in Medieval French}
    
    \vspace{2.6em}
    
    \small 
    \begin{tabular}{p{0.35\textwidth}p{0.6\textwidth}}
        \hline
        \textbf{Characteristic} & \textbf{Examples}  \\ 
        \midrule
        \midrule
        Spelling variation & assi, assy (engl. 'so') \\ 
        Medieval spelling & quando (engl. 'when'), qual (engl. 'which') instead of cuando, cual\\
        Medieval forms & fazer instead of modern hacer (engl. 'to do') \\
        \bottomrule
    \end{tabular}
    \caption*{(e) Characteristics and examples of linguistic challenges in Medieval Spanish}
\end{minipage}
\caption{Romance languages in medieval texts: geographical distribution of (a) French and Occitan varieties and (b) Spanish varieties, with characteristic linguistic challenges in (c) Medieval Occitan, (d) Medieval French, and (e) Medieval Spanish.}
\label{fig:combined}
\end{figure*}

Language technologies have made remarkable advances in recent years, particularly with the development of large language models \citep[LLMs; ][]{achiam2023gpt,grattafiori2024llama,guo2025deepseek} that achieve near-human performance on various natural language processing tasks for high-resource languages \citep{brown2020language, raffel2020exploring}. However, historical languages—especially those with limited extant corpora—remain largely excluded from this progress, creating a widening technological divide between modern and historical text processing capabilities \citep{piotrowski2012natural}. Part-of-speech (POS) tagging, which involves assigning grammatical categories to words in context, serves as a critical foundation for downstream processing of historical texts. This fundamental task enables higher-level linguistic analyses, facilitates information extraction, and supports the development of search tools crucial for humanities scholars exploring literary, cultural, and historical patterns \citep{bollmann2019large}. Medieval language varieties such as Medieval Occitan, French, and Spanish (cf. Figure \ref{fig:combined}) present particularly challenging cases for computational processing due to three interconnected factors that conventional NLP approaches fail to address effectively. First, these texts exhibit extreme orthographic inconsistency, with the same word appearing in multiple spellings even within the same document. Second, they display substantial dialectal variation across regions and time periods, creating a fragmented linguistic landscape. Third, they suffer from a severe scarcity of annotated resources, with some varieties having fewer than 10,000 annotated tokens available for model development \citep{zampieri2019natural}. Despite their immense cultural and historical significance—preserving literary traditions, scientific knowledge, and historical records of medieval Europe—these languages have received comparatively little attention in the computational linguistics community.
This technological gap not only impedes scholarly progress but also risks the gradual erosion of accessibility to important cultural heritage contained in these texts. This study addresses this critical research gap by conducting a comprehensive analysis of the factors that influence POS tagging performance in low-resource medieval languages when using modern open-source LLMs.

Our investigation is guided by the following research questions:
\begin{enumerate}
\item \textbf{RQ1:} How do the choice of prompting and decoding strategies affect the POS tagging accuracy in orthographically inconsistent texts? (investigated in Sections \ref{prompting} and \ref{decoding_strat})
\item \textbf{RQ2:} What fine-tuning strategies are most effective for adapting general-purpose LLMs to the specific challenges of historical language varieties? (discussed in Section \ref{finetuning})
\item \textbf{RQ3:} Can cross-lingual transfer learning effectively leverage knowledge from better-resourced historical languages to improve performance on extremely low-resource varieties? (analyzed in Section \ref{cltl})
\item \textbf{RQ4:} How do architectural choices in LLMs affect POS tagging performance across different medieval language varieties? (discussed in Section \ref{modelsize})
\end{enumerate}
\clearpage
We summarize our contributions as follows:
\begin{enumerate}
\item We have annotated and publicly released two Medieval Occitan datasets encompassing 135,667 tokens across various domains—medical, biblical, and dietary—to enhance further research in the field.
\item We conduct extensive experiments across seven datasets of three medieval Romance languages, systematically analyzing how model size, prompting strategies, decoding approaches, fine-tuning methods, and cross-lingual transfer learning affect POS tagging accuracy in orthographically inconsistent, dialectally diverse texts.
\item Based on our results and error analysis, we provide evidence-based recommendations for practitioners and researchers, designing a comprehensive strategy for enhancing POS tagging accuracy in similar low-resource historical language contexts.
\item We make our complete codebase, experimental results, and detailed error analysis publicly available\footnote{\url{https://github.com/YecanLee/POS_Low_Resource_Medieval_Languages}}.
\end{enumerate}

\section{Related work}
\label{sec:related_work}


\subsection{Methodological approaches for Low-resource settings}

In low-resource POS tagging, methodological appropriateness is fundamental. \citet{terhoeve2022highresourcemethodologicalbiaslowresource} identified critical training-data bias when high-resource techniques are applied without necessary adaptations, while \citet{bansal2021lowlowcomputationalperspective} proposed a nuanced computational framework for assessing language resource levels beyond simplistic data volume metrics. Although unsupervised and weakly supervised approaches offer potential solutions \citep{cardenas2019groundedunsuperviseduniversalpartofspeech}, they often perform poorly on truly low-resource languages without additional adaptations \citep{kann2020weaklysupervisedpostaggers}. Integrating traditional lexical resources with modern techniques has shown promise for improving performance in resource-constrained environments \citep{plank2018bestworldslexicalresources}.

\subsection{Cross-Lingual transfer and historical romance languages}

Cross-lingual transfer has emerged as promising for medieval Romance languages, leveraging their etymological relationship with better-resourced contemporary variants \citep{ponti2020parameterspacefactorizationzeroshot, chopra2024zeroresourcecrosslingualspeech}. \citet{aepli2022improvingzeroshotcrosslingualtransfer} demonstrated that character-level noise injection during training improves transfer for languages with shared roots—particularly relevant for orthographically variable medieval texts. However, \citet{vandenbulcke2024recipezeroshotpostagging} questioned these approaches' practical utility in scenarios with highly divergent historical variants. For Romance languages specifically, specialized resources include models for Medieval French POS tagging \citep{camps2021corpusmodelslemmatisationpostagging}, D'AlemBERT for Early Modern French \citep{gabay2022freemdalembertlargecorpus}, and surveys of approaches for Medieval Latin texts \citep{nowak2024efontesspeechtagginglemmatization}. Notably, \citet{schoeffel2025modernmodelsmedievaltexts} examined how prompting strategies, model scale, and language support affect performance on Medieval Occitan corpora, revealing critical limitations under extreme orthographic and syntactic variability.

\subsection{Determinant factors in tagging accuracy}

Multiple factors influence POS tagging accuracy in low-resource contexts. Contextual information becomes more critical in extreme low-resource settings \citep{edman2022importancecontextlowresource}, while tokenization strategies significantly impact performance \citep{blaschke2023doesmanipulatingtokenizationaid}—especially for non-standardized historical texts. Resource augmentation strategies include efficient use of limited annotation budgets through targeted active learning \citep{chaudhary2020reducingconfusionactivelearning} and leveraging external knowledge sources \citep{plank2018distantsupervisiondisparatesources, cao2019lowresourcetagginglearnedweakly}. Graph-based approaches have shown promise for multilingual label propagation \citep{imani2022graphbasedmultilinguallabelpropagation}. Architecture innovations include character-aware hierarchical transformers \citep{riemenschneider2024heidelbergbostonsigtyp2024}, token-level prompt decomposition \citep{ma2024toprotokenlevelpromptdecomposition}, and hybrid approaches combining pre-trained language models with hand-crafted features \citep{zhou2022bridgingpretrainedlanguagemodels}. For endangered Romance variants, linguistic affinities can be effectively leveraged \citep{anastasopoulos2018partofspeechtaggingendangeredlanguage}, while predicted universal POS tags can support parsing even in severely resource-constrained settings \citep{anderson2021faltapanbuenasson}.

\section{Experimental setup}
\label{experimental_setup}

We evaluate seven open-source instruction-tuned models (2B-14B parameters) on POS tagging for seven datasets of three medieval Romance languages. Our experiments include: (1) prompting with various strategies to assess zero-shot and few-shot performance, and (2) fine-tuning to investigate both single-dataset training and cross-language transfer effects. We maintain consistent evaluation protocols across all experiments. Table \ref{tab:experimental_setup} summarizes our experimental setup, with additional details on prompting strategies available in Table \ref{tab:prompting_strategies}, Appendix \ref{a:prompting_strategies}.

\subsection{Data}
\label{Data}

\paragraph{Medieval Occitan.} We utilize three Medieval Occitan texts. The first, British Library's Harley 7403 manuscript, contains several texts including \textit{l'Évangile de Nicodème} translation. Building on \cite{Wiedner2023}'s work, it was digitized, transcribed using an HTR model for Medieval Occitan handwriting, and manually revised. 

The second text, Nouvelle Acquisition Française 6195 (NAF6195), a 14th-century Provençal manuscript also known as manuscript M of the \textit{Vida de Sant Honorat}, was digitized at the Bibliothèque Nationale de France. Its semi-automatic transcription was initially annotated using a modern Occitan POS tagger \citep{PoujadeInProgress} before manual corrections. This previously unused version contains graphical variants distinct from the DOM dictionary\footnote{\href{The Dictionaire de l'occitan médiéval (DOM)}{\url{https://dom-en-ligne.de/}} is the reference dictionary for Medieval Occitan with 79,913 entries, 38,869 unique lemmas, and 41,044 graphical variants as of March 2025.} and existing editions.

The third text, \textit{On Surgery and Instruments} by Abū l-Qāsim al-Ḥalaf al-Zahrāwī (Albucasis), is based on the Bibliothèque de l'Université (Montpellier) manuscript, edited by P.T. Ricketts, converted to TEI by D. Billy, and released under CC BY-NC-SA 4.0. This 14th-century translation of the Arabic medical encyclopedia al-Tasrif features specialized vocabulary in surgery, anatomy, pharmacy, botany, and zoology. Figure \ref{fig:combined}(c) illustrates Medieval Occitan linguistic specialties.

\begin{figure}[H]
\centering
\includegraphics[width=0.9\textwidth]{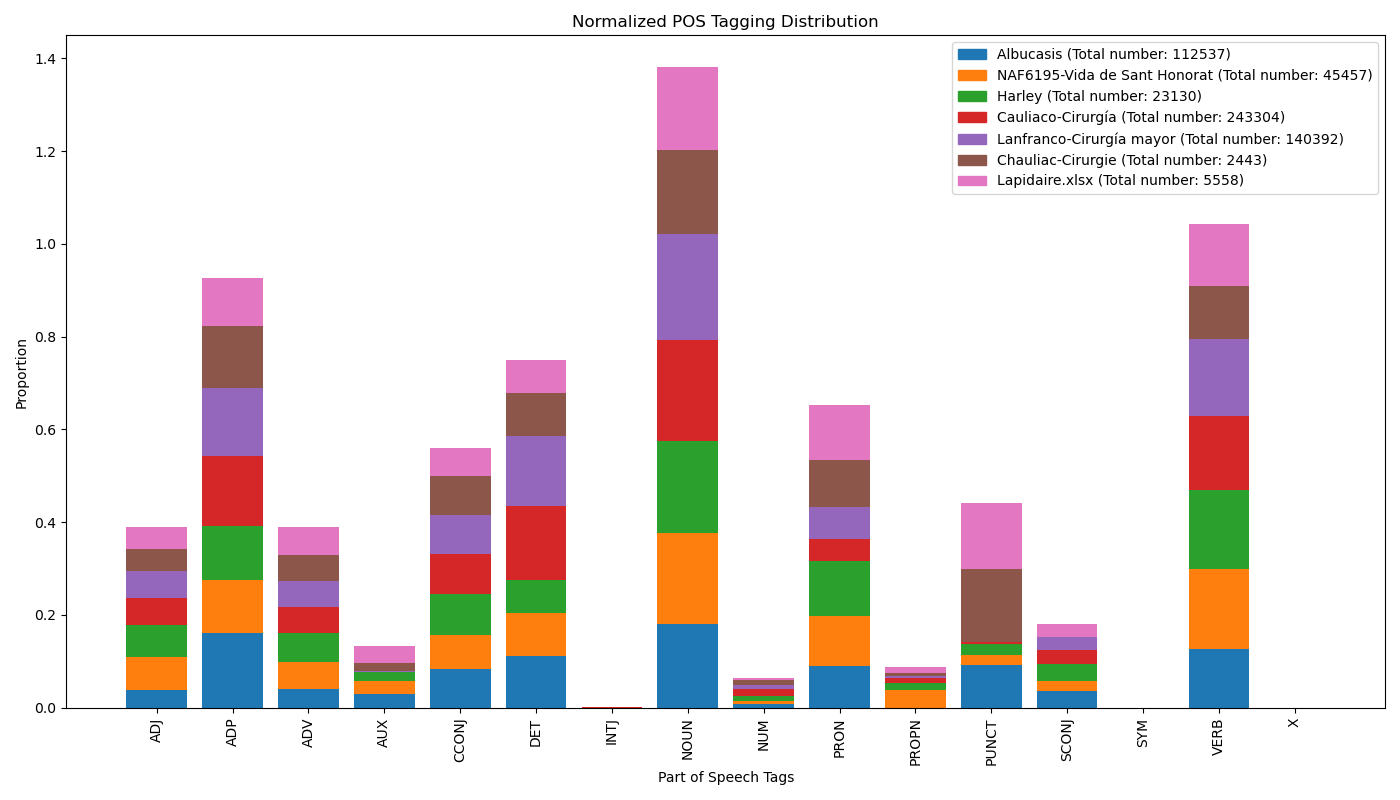}
\caption{Normalized Part-of-Speech distribution for the corpus: Albucasis (blue), Vida de Sant Honorat (orange), Harley (green), Cauliaco (red), Lanfranco (purple), Chauliac (brown), and Lapidiare (pink).}
\label{fig:figure_two}
\end{figure}

\paragraph{Medieval French and Medieval Spanish.} We include two Medieval French and two Medieval Spanish texts, primarily medical in nature. For Medieval French, we selected: (1) a 12th-century Anglo-Norman lapidarium \citep{Lapidaire}, accessible via \citet{prevost:hal-04681591}; and (2) \textit{Anathomie}, from Gui de Chauliac's \textit{Grande Chirurgie} (15th century), based on Montpellier's Bibliothèque de la Faculté de Médecine manuscript H 184, with POS tagging provided by \cite{alma991083835439705501}. Figure \ref{fig:combined}(d) illustrates these texts' features.

The Spanish texts are: (1) \textit{Compendio de cirugía}, an Medieval Spanish translation of Lanfranco de Milán's \textit{Liber de chirurgia}, from a 1481 manuscript (Madrid, Nacional MSS/2147), digitized and annotated by the Hispanic Seminary \citep{Jover2011a}; and (2) the Spanish translation of Gui de Chauliac's \textit{Chirurgia Magna}, an incunable from 1498 (Madrid, Nacional INC/196), transcribed by María Teresa Herrera \citep{Jover2011b}. Figure \ref{fig:combined}(e) summarizes these texts' specific traits. Figure \ref{fig:figure_two} shows the normalized distribution of POS tags, following the Universal Dependency Standard\footnote{The tag set is based on \url{https://universaldependencies.org/u/pos/}.}.

\begin{table}[htbp]
\centering
\small
\renewcommand{\arraystretch}{1.2}
\definecolor{headercolor}{RGB}{240,240,240}

\begin{tabular}{p{3cm}p{9cm}}
\toprule
\rowcolor{headercolor} \multicolumn{2}{c}{\textbf{Models}} \\
\midrule
\textbf{Open-Source Models} & 
• Phi4-14B \citep{abdin2024phi4technicalreport}\\
& • Qwen2.5-14B \citep{qwen2025qwen25technicalreport}\\
& • Gemma2-9B \citep{gemmateam2024gemma2improvingopen}\\
& • Llama-3.1-8B \citep{grattafiori2024llama3herdmodels}\\
& • Aya-8B \citep{aryabumi2024aya23openweight}\\
& • Mistral-7B-Instruct-v0.2 \citep{jiang2023mistral}\\
& • Gemma2-2B \citep{gemmateam2024gemma2improvingopen}\\
& \\
& Language supported are documented in Table \ref{tab:supported_languages},  Appendix \ref{a:supported_languages}\\
\midrule
\rowcolor{headercolor} \multicolumn{2}{c}{\textbf{Datasets}} \\
\midrule
\textbf{Medieval Occitan} & NAF, Harley, Albucasis \\
\textbf{Medieval French} & Chauliac, Lapidaire \\
\textbf{Medieval Spanish} & Cauliaco, Lanfranco \\
\midrule
\rowcolor{headercolor} \multicolumn{2}{c}{\textbf{Task 1: Prompting (24 Experiments)}} \\
\midrule
\textbf{Models Used} & Gemma2-9B \\
\textbf{Prompt Types} & Zero-shot, Few-shot (cf. Appendix \ref{a:prompting_strategies}, Table \ref{tab:prompting_strategies}) \\
\textbf{Decoding Strategies} & 
• Greedy search \citep{Freitag_2017} \\
& • Temperature sampling \citep{ackley1985learning} with $T = 0.3$\\
& • Temperature sampling \citep{ackley1985learning} with $T = 0.9$\\
& • Top-$p$ sampling \citep{holtzman2019curious}  with $p = 0.95$\\
\textbf{Datasets Used} & NAF, Chauliac, Lanfranco\\
\textbf{Evaluation} & • Based on LLM-generated JSON files \\
\midrule
\rowcolor{headercolor} \multicolumn{2}{c}{\textbf{Task 2: Fine-tuning (98 Experiments)}} \\
\midrule
\textbf{Models Used} & All seven models \\
\textbf{Datasets Used} & All seven datasets \\
\textbf{Evaluation} & 
\textbf{Setting (a):} Single-dataset fine-tuning (1-to-1)\\
& • Fine-tune each model on 80\% of a single dataset\\
& • Evaluate on 20\% of the same dataset\\
\cmidrule{2-2}
& \textbf{Setting (b):} Cross-language transfer (N-to-1)\\
& • Fine-tune each model on 80\% of all datasets combined\\
& • Evaluate on 20\% of a specific target dataset\\
\bottomrule
\end{tabular}
\caption{Experimental setup for POS tagging of medieval romance languages. For the evaluation and analysis, we focused on accuracy, but we make available measurements of precision, recall, and F1-measures. A comprehensive description of these metrics and their formal definitions is provided in Appendix \ref{a:metrics}. All experiments were conducted using an NVIDIA RTX 4090-24GB GPU for the 7B-12B models and NVIDIA H100-96GB GPU for the 14B models.}
\label{tab:experimental_setup}
\end{table}

\section{Results}
\label{results}

\subsection{Task 1: Prompting}

To illustrate the impact of various prompting and decoding strategies (see Table \ref{tab:pos-tagging-comparison}), we focus our analysis on the Gemma2-9B model. We evaluate this model using zero-shot and few-shot prompting approaches across three datasets, employing four distinct decoding strategies. This focused presentation allows for clearer visualization of effects, as we observed similar patterns across multiple models and datasets. For comprehensive results across all  configurations, we refer readers to our GitHub repository\footnote{\url{https://github.com/YecanLee/POS_Low_Resource_Medieval_Languages}}, and to our supplementary class-specific analysis in Section \ref{a:further_analysis_pos_class}.

\begin{table}[!ht]
\centering
\resizebox{\textwidth}{!}{%
\begin{tabular}{l|ccc|ccc|ccc|ccc}
\toprule
\multirow{2}{*}{\textbf{Model (Decoding)}} & \multicolumn{3}{c|}{\textbf{NAF}} & \multicolumn{3}{c|}{\textbf{Chauliac}} & \multicolumn{3}{c|}{\textbf{Lanfranco}} & \multicolumn{3}{c}{\textbf{Average}} \\
\cmidrule(lr){2-4} \cmidrule(lr){5-7} \cmidrule(lr){8-10} \cmidrule(lr){11-13}
 & \textbf{Zero-shot} & \textbf{Few-shot} & \textbf{Delta} & \textbf{Zero-shot} & \textbf{Few-shot} & \textbf{Delta} & \textbf{Zero-shot} & \textbf{Few-shot} & \textbf{Delta} & \textbf{Zero-shot} & \textbf{Few-shot} & \textbf{Delta} \\
\midrule
gemma\_9b (greedy) & 0.7028 & 0.7042 & +0.0014 & \cellcolor[HTML]{C6EFCE}\textbf{0.8627} & \cellcolor[HTML]{C6EFCE}\textbf{0.8815} & +0.0188 & \cellcolor[HTML]{FFC7CE}\textbf{0.7632} & \cellcolor[HTML]{FFC7CE}\textbf{0.7791} & \cellcolor[HTML]{FFC7CE}\textbf{+0.0159} & 0.7762 & 0.7883 & +0.0120 \\
gemma\_9b (temp = 0.3) & 0.6952 & 0.6957 & \cellcolor[HTML]{FFC7CE}\textbf{+0.0005} & \cellcolor[HTML]{FFC7CE}\textbf{0.8217} & 0.8580 & \cellcolor[HTML]{C6EFCE}\textbf{+0.0363} & \cellcolor[HTML]{C6EFCE}\textbf{0.8003} & \cellcolor[HTML]{C6EFCE}\textbf{0.8169} & \cellcolor[HTML]{C6EFCE}\textbf{+0.0166} & 0.7724 & \cellcolor[HTML]{C6EFCE}\textbf{0.7902} & \cellcolor[HTML]{C6EFCE}\textbf{+0.0178} \\
gemma\_9b (temp = 0.9) & \cellcolor[HTML]{C6EFCE}\textbf{0.7062} & \cellcolor[HTML]{C6EFCE}\textbf{0.7085} & \cellcolor[HTML]{C6EFCE}\textbf{+0.0023} & 0.8473 & \cellcolor[HTML]{FFC7CE}\textbf{0.8473} & \cellcolor[HTML]{FFC7CE}\textbf{+0.0000} & 0.7885 & 0.8050 & 0.0165 & \cellcolor[HTML]{C6EFCE}\textbf{0.7807} & 0.7869 & \cellcolor[HTML]{FFC7CE}\textbf{+0.0063} \\
gemma\_9b (p = 0.95) & \cellcolor[HTML]{FFC7CE}\textbf{0.6877} & \cellcolor[HTML]{FFC7CE}\textbf{0.6891} & +0.0014 & 0.8368 & 0.8553 & +0.0185 & 0.7814 & 0.7977 & 0.0163 & \cellcolor[HTML]{FFC7CE}\textbf{0.7686} & \cellcolor[HTML]{FFC7CE}\textbf{0.7807} & +0.0120 \\
\midrule
\textbf{Average} & 0.6980 & 0.6994 & +0.0014 & 0.8421 & 0.8605 & +0.0184 & 0.7833 & 0.7997 & +0.0163 & 0.7745 & 0.7865 & +0.0120 \\
\bottomrule
\end{tabular}%
}
\caption{POS tagging accuracy comparison between Zero-shot and Few-shot prompting strategies across different datasets and decoding methods. positive Delta values represent improvements through Few-shot. Highest values per column are highlighted in \colorbox[HTML]{C6EFCE}{\textbf{green}}, lowest in \colorbox[HTML]{FFC7CE}{\textbf{red}}.}
\label{tab:pos-tagging-comparison}
\end{table}

\subsubsection{Effect of prompting}
\label{prompting}

The comparison between Zero-shot and Few-shot prompting strategies reveals a consistent pattern of improvement across all datasets and decoding strategies. Few-shot prompting consistently outperforms Zero-shot prompting, with an average accuracy improvement of +0.0120 across all configurations. This improvement, while modest, demonstrates that providing examples during prompting helps the model better understand the POS tagging task.
The magnitude of improvement varies across datasets and decoding strategies. The Chauliac dataset shows the largest average improvement (+0.0184), followed by Lanfranco (+0.0163) and NAF (+0.0014). This suggests that the benefit of Few-shot prompting may depend on the linguistic characteristics of the dataset, with medieval medical texts like Chauliac potentially benefiting more from explicit examples.
The most substantial improvement was observed with temperature = 0.3 on the Chauliac dataset (+0.0363), while temperature = 0.9 showed minimal improvement on the same dataset (+0.0000). This indicates that the effectiveness of Few-shot prompting interacts with the decoding strategy, with more deterministic decoding approaches potentially benefiting more from example-based guidance.

\subsubsection{Effect of decoding strategies}
\label{decoding_strat}

The analysis reveals that different decoding strategies perform optimally depending on the dataset and prompting approach. Temperature = 0.9 achieves the highest Zero-shot accuracy on average (0.7807), while temperature = 0.3 produces the best Few-shot results (0.7902). This suggests that higher randomness (temperature = 0.9) might help the model explore more possible POS tags when working without examples, while more deterministic decoding (temperature = 0.3) works better when guided by examples.
Dataset-specific patterns are also evident. For the NAF dataset, temperature = 0.9 consistently performs best across both Zero-shot and Few-shot approaches. The Chauliac dataset benefits most from greedy decoding, achieving the highest accuracy in both prompting scenarios (0.8627 and 0.8815). For Lanfranco, temperature = 0.3 proves most effective (0.8003 and 0.8169). These variations suggest that optimal decoding strategies may depend on the linguistic features and complexity of each dataset.
The nucleus sampling approach (p = 0.95) generally underperforms compared to other strategies, showing the lowest average accuracy in both Zero-shot (0.7686) and Few-shot (0.7807) scenarios. This indicates that for the structured task of POS tagging, more focused decoding approaches may be preferable to those that sample from a wider distribution of possibilities.

\subsection{Task 2: Fine-tuning}

We evaluated seven open-source language models on seven medieval romance language datasets to assess their part-of-speech (POS) tagging capabilities. Table~\ref{tab:fine-tuning} presents the accuracy scores achieved by each model when fine-tuned and evaluated on individual datasets.

\begin{table}[!ht]
\centering
\resizebox{\textwidth}{!}{%
\begin{tabular}{lccc|cc|cc|c}
\toprule
\multirow{2}{*}{\textbf{Model}} & \multicolumn{3}{c|}{\textbf{Medieval Occitan}} & \multicolumn{2}{c|}{\textbf{Medieval French}} & \multicolumn{2}{c|}{\textbf{Medieval Spanish}} & \multirow{2}{*}{\textbf{Average}} \\
\cmidrule(lr){2-4} \cmidrule(lr){5-6} \cmidrule(lr){7-8}
 & \textbf{NAF} & \textbf{Harley} & \textbf{Albucasis} & \textbf{Chauliac} & \textbf{Lapidaire} & \textbf{Cauliaco} & \textbf{Lanfranco} & \\
\midrule
phi4\_14b & 0.7892 & 0.6970 & 0.9054 & 0.8115 & \cellcolor[HTML]{FFC7CE}\textbf{0.8356} & \cellcolor[HTML]{FFC7CE}\textbf{0.8691} & 0.8642 & 0.8246 \\
qwen\_14b & 0.7702 & 0.6804 & 0.9093 & \cellcolor[HTML]{FFC7CE}\textbf{0.8036} & 0.8418 & 0.8707 & 0.8614 & 0.8196 \\
gemma\_9b & 0.8057 & 0.7054 & 0.9153 & 0.8413 & 0.8717 & 0.8974 & \cellcolor[HTML]{C6EFCE}\textbf{0.9032} & 0.8486 \\
aya\_8b & \cellcolor[HTML]{C6EFCE}\textbf{0.8159} & \cellcolor[HTML]{C6EFCE}\textbf{0.7606} & \cellcolor[HTML]{C6EFCE}\textbf{0.9173} & \cellcolor[HTML]{C6EFCE}\textbf{0.8433} & \cellcolor[HTML]{C6EFCE}\textbf{0.8755} & \cellcolor[HTML]{C6EFCE}\textbf{0.9025} & 0.8930 & \cellcolor[HTML]{C6EFCE}\textbf{0.8583} \\
llama3\_1\_8b & 0.7799 & 0.6876 & 0.9078 & 0.8115 & 0.8418 & 0.8701 & 0.9013 & 0.8286 \\
mistral\_7b & \cellcolor[HTML]{FFC7CE}\textbf{0.6354} & \cellcolor[HTML]{FFC7CE}\textbf{0.6429} & \cellcolor[HTML]{FFC7CE}\textbf{0.8962} & 0.8095 & 0.8742 & 0.8780 & \cellcolor[HTML]{FFC7CE}\textbf{0.7532} & \cellcolor[HTML]{FFC7CE}\textbf{0.7842} \\
gemma\_2b & 0.8011 & 0.7025 & 0.9092 & 0.8115 & 0.8406 & 0.8826 & 0.9009 & 0.8355 \\
\midrule
\textbf{Average} & 0.7711 & 0.6966 & 0.9086 & 0.8189 & 0.8545 & 0.8815 & 0.8682 & 0.8285 \\
\bottomrule
\end{tabular}%
}
\caption{POS tagging accuracy for individual models fine-tuned on specific datasets. The highest scores per dataset are highlighted in \colorbox[HTML]{C6EFCE}{\textbf{green}}, while lowest scores are highlighted in \colorbox[HTML]{FFC7CE}{\textbf{red}}.}
\label{tab:fine-tuning}
\end{table}

Table~\ref{tab:multilingual} shows the results when adopting a multilingual transfer learning approach, where models were trained on all datasets simultaneously and then evaluated on each individual dataset.

\begin{table}[!ht]
\centering
\resizebox{\textwidth}{!}{%
\begin{tabular}{lccc|cc|cc|c}
\toprule
\multirow{2}{*}{\textbf{Model}} & \multicolumn{3}{c|}{\textbf{Medieval Occitan}} & \multicolumn{2}{c|}{\textbf{Medieval French}} & \multicolumn{2}{c|}{\textbf{Medieval Spanish}} & \multirow{2}{*}{\textbf{Average}} \\
\cmidrule(lr){2-4} \cmidrule(lr){5-6} \cmidrule(lr){7-8}
 & \textbf{NAF} & \textbf{Harley} & \textbf{Albucasis} & \textbf{Chauliac} & \textbf{Lapidaire} & \textbf{Cauliaco} & \textbf{Lanfranco} & \\
\midrule
phi4\_14b & 0.7731 & 0.7424 & \cellcolor[HTML]{FFC7CE}\textbf{0.9008} & 0.8552 & 0.8269 & 0.8686 & \cellcolor[HTML]{FFC7CE}\textbf{0.8494} & 0.8309 \\
qwen\_14b & \cellcolor[HTML]{FFC7CE}\textbf{0.7711} & 0.7451 & 0.9063 & 0.8552 & \cellcolor[HTML]{C6EFCE}\textbf{0.8406} & \cellcolor[HTML]{FFC7CE}\textbf{0.8592} & 0.8494 & 0.8324 \\
gemma\_9b & \cellcolor[HTML]{C6EFCE}\textbf{0.8086} & \cellcolor[HTML]{C6EFCE}\textbf{0.7813} & 0.9116 & \cellcolor[HTML]{C6EFCE}\textbf{0.8631} & 0.8219 & 0.8914 & 0.8896 & 0.8525 \\
aya\_8b & 0.8010 & 0.7797 & \cellcolor[HTML]{C6EFCE}\textbf{0.9123} & 0.8333 & 0.8381 & \cellcolor[HTML]{C6EFCE}\textbf{0.9049} & \cellcolor[HTML]{C6EFCE}\textbf{0.8999} & \cellcolor[HTML]{C6EFCE}\textbf{0.8527} \\
llama3\_1\_8b & 0.7752 & \cellcolor[HTML]{FFC7CE}\textbf{0.7305} & 0.9064 & \cellcolor[HTML]{FFC7CE}\textbf{0.8194} & 0.8219 & 0.8740 & 0.8673 & 0.8278 \\
mistral\_7b & 0.7780 & 0.7370 & 0.9065 & 0.8274 & \cellcolor[HTML]{FFC7CE}\textbf{0.7597} & 0.8776 & 0.8762 & \cellcolor[HTML]{FFC7CE}\textbf{0.8232} \\
gemma\_2b & 0.7801 & 0.7586 & 0.9057 & 0.8611 & 0.8070 & 0.8884 & 0.8779 & 0.8398 \\
\midrule
\textbf{Average} & 0.7839 & 0.7535 & 0.9071 & 0.8450 & 0.8166 & 0.8806 & 0.8728 & 0.8370 \\
\bottomrule
\end{tabular}%
}
\caption{POS tagging accuracy for models trained with multilingual transfer learning. The highest scores per dataset are highlighted in \colorbox[HTML]{C6EFCE}{\textbf{green}}, while lowest scores are highlighted in \colorbox[HTML]{FFC7CE}{\textbf{red}}.}
\label{tab:multilingual}
\end{table}

\subsubsection{Effect of fine-tuning}
\label{finetuning}

Fine-tuning substantially enhances POS tagging performance compared to prompting-based approaches. For Gemma2-9B, we observe remarkable improvements across all three comparable datasets. On the NAF dataset, fine-tuning achieves an accuracy of 0.8057, dramatically outperforming the best prompting-based result of 0.7085 (few-shot with temperature = 0.9) by +0.0972 absolute points. For the Chauliac dataset, fine-tuning reaches 0.8413 accuracy, compared to 0.8815 under few-shot prompting with greedy decoding—a slight decrease of 0.0402. Most notably, on the Lanfranco dataset, fine-tuning delivers 0.9032 accuracy versus the best prompting result of 0.8169 (few-shot with temperature = 0.3), representing a substantial improvement of +0.0863.

These results indicate that while few-shot prompting with optimized decoding strategies can achieve reasonable performance—particularly for Medieval French texts like Chauliac—fine-tuning consistently provides more robust performance across language varieties. The efficiency gap is particularly pronounced for Medieval Occitan and Spanish, where fine-tuning appears to better capture the orthographic inconsistencies and dialectal variations characteristic of these languages. This suggests that the representations learned during fine-tuning more effectively encode the complex linguistic patterns of medieval texts compared to the in-context learning facilitated by few-shot prompting.

\subsubsection{Effect of cross-lingual transfer learning}
\label{cltl}

To directly compare the benefits of multilingual transfer learning over individual fine-tuning, we computed the absolute improvements in accuracy for each model-dataset combination, see Table~\ref{tab:improvement}.

\begin{table}[H]
\centering
\resizebox{\textwidth}{!}{%
\begin{tabular}{lccc|cc|cc|c}
\toprule
\multirow{2}{*}{\textbf{Model}} & \multicolumn{3}{c|}{\textbf{Medieval Occitan}} & \multicolumn{2}{c|}{\textbf{Medieval French}} & \multicolumn{2}{c|}{\textbf{Medieval Spanish}} & \multirow{2}{*}{\textbf{Average}} \\
\cmidrule(lr){2-4} \cmidrule(lr){5-6} \cmidrule(lr){7-8}
 & \textbf{NAF} & \textbf{Harley} & \textbf{Albucasis} & \textbf{Chauliac} & \textbf{Lapidaire} & \textbf{Cauliaco} & \textbf{Lanfranco} & \\
\midrule
phi4\_14b & -1.61 & +4.54 & -0.46 & +4.37 & -0.87 & -0.05 & -1.48 & +0.63 \\
qwen\_14b & +0.09 & +6.47 & -0.30 & \cellcolor[HTML]{C6EFCE}\textbf{+5.16} & \cellcolor[HTML]{C6EFCE}\textbf{-0.12} & \cellcolor[HTML]{FFC7CE}\textbf{-1.15} & -1.20 & +1.28 \\
gemma\_9b & +0.29 & +7.59 & -0.37 & +2.18 & -4.98 & -0.60 & -1.36 & +0.39 \\
aya\_8b & -1.49 & \cellcolor[HTML]{FFC7CE}\textbf{+1.91} & \cellcolor[HTML]{FFC7CE}\textbf{-0.50} & \cellcolor[HTML]{FFC7CE}\textbf{-1.00} & -3.74 & +0.24 & +0.69 & \cellcolor[HTML]{FFC7CE}\textbf{-0.56} \\
llama3\_1\_8b & -0.47 & +4.29 & -0.14 & +0.79 & -1.99 & +0.39 & \cellcolor[HTML]{FFC7CE}\textbf{-3.40} & -0.08 \\
mistral\_7b & \cellcolor[HTML]{C6EFCE}\textbf{+14.26} & \cellcolor[HTML]{C6EFCE}\textbf{+9.41} & \cellcolor[HTML]{C6EFCE}\textbf{+1.03} & +1.79 & \cellcolor[HTML]{FFC7CE}\textbf{-11.45} & -0.04 & \cellcolor[HTML]{C6EFCE}\textbf{+12.30} & \cellcolor[HTML]{C6EFCE}\textbf{+3.90} \\
gemma\_2b & \cellcolor[HTML]{FFC7CE}\textbf{-2.10} & +5.61 & -0.35 & +4.96 & -3.36 & \cellcolor[HTML]{C6EFCE}\textbf{+0.58} & -2.30 & +0.43 \\
\midrule
\textbf{Average} & +1.28 & +5.69 & -0.16 & +2.61 & -3.79 & -0.09 & +0.46 & +0.86 \\
\bottomrule
\end{tabular}%
}
\caption{Absolute improvement as percentual points in POS tagging accuracy when using multilingual transfer learning compared to individual fine-tuning. Positive values indicate performance gains. The highest scores per dataset are highlighted in \colorbox[HTML]{C6EFCE}{\textbf{green}}, while lowest scores are highlighted in \colorbox[HTML]{FFC7CE}{\textbf{red}}.}
\label{tab:improvement}
\end{table}

Cross-lingual transfer learning via multilingual fine-tuning yields an average improvement of +0.86 percentage points over single-dataset training, with notably heterogeneous effects. Medieval Occitan benefits most significantly, particularly the lower-resource Harley corpus (+5.69 percentage points), suggesting that additional signals from related Romance languages help overcome data scarcity. Medieval French displays contrasting patterns: Chauliac improves substantially (+2.61 points) while Lapidaire deteriorates (-3.79 points), likely reflecting different levels of shared vocabulary with other corpus texts. Medieval Spanish shows minimal changes, suggesting these datasets may already contain sufficient training data. Model-specific responses reveal that Mistral-7B, primarily trained on English, gains dramatically from multilingual exposure (+14.26 on NAF, +9.41 on Harley), while Aya-8B, pre-trained on multiple Romance languages, shows minimal benefit (-0.56 average), indicating its representations already effectively capture cross-linguistic patterns.

\subsubsection{Effect of model size}
\label{modelsize}

Model size alone does not predict POS tagging performance for medieval languages. The 8B-parameter Aya model consistently outperforms larger 14B-parameter models in both single-dataset (0.8583 average accuracy vs. 0.8246 for Phi4-14B) and multilingual fine-tuning scenarios (0.8527 vs. 0.8309). Similarly, Gemma2-2B (2B parameters) delivers performance (0.8355) exceeding both 14B-parameter models. These findings suggest that pre-training quality, architecture efficiency, and language coverage substantially outweigh raw parameter count. Models with explicit Romance language pre-training demonstrate superior performance even with fewer parameters. The poor performance of Mistral-7B (0.7842) despite its size further emphasizes that model size alone is insufficient for historical language processing.

\section{Practical recommendations}
\label{recommendations}

Based on our experiments, we recommend the following strategies for POS tagging medieval Romance languages:

\textbf{Model selection:}
\begin{itemize}
    \item Prioritize models pre-trained on Romance languages (Aya-8B, Phi4-14B) over larger monolingual alternatives
    \item Consider Gemma2-2B for resource-constrained environments, as it rivals larger models at lower computational cost
    \item Avoid models lacking Romance language pre-training, regardless of parameter count
\end{itemize}

\textbf{Optimization approach:}
\begin{itemize}
    \item When annotated data is available, even in limited quantities, always prefer fine-tuning over prompting
    \item For low-resource varieties (particularly Occitan), pool training data from multiple medieval Romance languages
    \item For medical texts, leverage cross-lingual transfer; for specialized domains (e.g., lapidaries), evaluate potential negative transfer effects first
    \item When fine-tuning is infeasible, use few-shot prompting with language-specific decoding parameters: temperature=0.9 for Occitan, greedy decoding for French, temperature=0.3 for Spanish
\end{itemize}

\textbf{Deployment workflow:}
Begin with Aya-8B or Gemma2-2B, fine-tune on pooled medieval Romance data when target language data is limited, implement post-processing rules for frequent tokens, and document language-specific patterns identified during error analysis.

\section{Conclusion}
\label{conclusion}

This study provides a comprehensive evaluation of part-of-speech tagging for medieval Romance languages using modern large language models. Through rigorous experimentation across seven datasets representing Medieval Occitan, French, and Spanish varieties, we have identified critical factors that determine tagging accuracy in these challenging low-resource, orthographically inconsistent historical texts. Our findings indicate that fine-tuning consistently outperforms prompting approaches, with multilingual transfer learning offering substantial benefits for extremely low-resource varieties. We have shown that model architecture and pre-training data composition substantially outweigh raw parameter count, with the 8B-parameter Aya model outperforming larger 14B-parameter alternatives through better Romance language representations. 
We have also documented language-specific patterns in how different prompting strategies, decoding approaches, and cross-lingual transfer techniques affect performance. These insights culminate in a set of practical recommendations that can guide practitioners in selecting appropriate models and optimization strategies for medieval language processing. By releasing two new Medieval Occitan datasets comprising 135,667 tokens, along with our extensive codebase and experimental results, we hope to contribute resources to the computational linguistics and digital humanities communities. This work addresses a critical technological gap in historical text processing and lays the groundwork for more accessible computational analysis of important cultural heritage materials preserved in medieval Romance languages.

\section{Limitations}
\label{limitations}

Despite our comprehensive evaluation, several limitations warrant consideration and point toward future research directions: First, our study focuses exclusively on Romance languages, leaving open questions about how our findings might generalize to medieval Germanic, Slavic, or non-European language families. Additionally, while we examined seven datasets, they primarily represent formal, literary, and medical texts—potentially limiting applicability to other genres such as administrative documents, personal correspondence, or vernacular literature. Moreover, our evaluation emphasizes overall accuracy, potentially obscuring performance variations across different POS categories. Future work should incorporate more fine-grained analysis of tag-specific metrics, particularly for syntactically ambiguous categories that are challenging in medieval texts.

Several promising research avenues emerge from this work:
\begin{itemize}
    \item Developing specialized tokenizers for medieval languages that better capture orthographic variation and morphological patterns.
    \item Investigating how other NLP tasks (e.g., named entity recognition, dependency parsing) can benefit from the POS tagging improvements demonstrated here.
    \item Extending our methodology to non-Romance medieval languages to assess the generalizability of our findings.
\end{itemize}

\section*{Acknowledgments}

Matthias Aßenmacher was funded by the Deutsche
Forschungsgemeinschaft (DFG, German Research
Foundation) under the National Research Data Infrastructure – NFDI 27/1 - 460037581. Additionally, we thank the Leibniz-Rechenzentrum der Bayerischen Akademie der Wissenschaften (LRZ) for
providing computational resources essential for this
research.

\section*{Ethics Statement}
This work involves the use of publicly available datasets and does not involve human
subjects or any personally identifiable information.
We declare that we have no conflicts of interest that
could potentially influence the outcomes, interpretations, or conclusions of this research. All funding
sources supporting this study are acknowledged in
the acknowledgments section. We have made our
best effort to document our methodology, experiments, and results accurately and are committed to
sharing our code, data, and other relevant resources
to foster reproducibility and further advancements
in research.

\bibliography{colm2025_conference}
\bibliographystyle{colm2025_conference}

\clearpage

\appendix
\section*{Appendix}

\section{Supported languages (pre-training)}
\label{a:supported_languages}

\begin{table}[H]
\centering
\resizebox{\textwidth}{!}{%
\begin{tabular}{lcccccccc}
\hline
\textbf{Model} & \textbf{Occitan} & \textbf{French} & \textbf{Spanish} & \textbf{Italian} & \textbf{Portuguese} & \textbf{Romanian} & \textbf{Arabic} & \textbf{English} \\
\hline
phi4\_14b      & \cmark & \cmark & \cmark & \cmark & \cmark & \cmark & \cmark & \cmark \\
qwen\_14b      &        & \cmark & \cmark & \cmark & \cmark &        & \cmark & \cmark \\
gemma\_9b      &        &        &        &        &        &        &        & \cmark \\
aya\_8b        &        & \cmark & \cmark & \cmark & \cmark & \cmark & \cmark & \cmark \\
llama3\_1\_8b  &        & \cmark & \cmark & \cmark & \cmark &        &        & \cmark \\
mistral\_7b    &        &        &        &        &        &        &        & \cmark \\
gemma\_2b      &        &        &        &        &        &        &        & \cmark \\
\hline
\end{tabular}
}
\caption{Overview of supported languages by each model. Note that none of the models supports Medieval Occitan, Medieval French, or Medieval Spanish.}
\label{tab:supported_languages}
\end{table}

\section{Prompting strategies}
\label{a:prompting_strategies}

\begin{table}[H]
    \centering
    \begin{tabular}{|l|p{10cm}|}
        \hline
        \textbf{Prompting Strategy} & \textbf{Prompt} \\
        \hline
        Zero-shot & \textit{You are a linguistic expert in Medieval Romance languages. 
            Please perform part-of-speech tagging  and assign Universal Dependencies Part-of-Speech (UD POS).
            Return the results as a JSON array of objects, each containing only the ‘word' and ‘upos' keys.
            Ensure that the JSON array is properly formatted and closed.
            The output must be only the JSON array without any additional text, explanations, or formatting.} \\
        \hline
        Few-shot & \textit{You are a medieval Romance language expert specializing in linguistic analysis.
        Your task is to analyze the given text and assign Universal Dependencies Part-of-Speech (UD POS) tags to each word.
        Return the results as a JSON array of objects, each containing only the 'word' and 'upos' keys.
        Ensure that the JSON array is properly formatted and closed.
        The output must be only the JSON array without any additional text, explanations, or formatting.
        The medieval Romance language have spelling variations. The syntax is not fully aligned with modern standards yet.
        Besides there are medieval forms that cannot be found in the corresponding modern language.
        Here are two examples of etymological twins: 
        tercia, tersa (Medieval Occitan), tierce (Medieval French), tercera (Medieval Spanish).
        sanguina, sanc (Medieval Occitan), sang (Medieval French), sangre (Medieval Spanish). 
        } \\
        \hline
        
    \end{tabular}
    \caption{Comparison of different prompting strategies for UD POS tagging.}
    \label{tab:prompting_strategies}
\end{table}

\section{Evaluation metrics}
\label{a:metrics}

We assessed our model using several standard metrics, defined as follows.

\paragraph{Accuracy} 
Accuracy quantifies the proportion of correctly predicted POS tags relative to the total number of tags:
\begin{equation}
    \text{Accuracy} = \frac{TP + TN}{TP + TN + FP + FN},
\end{equation}
where \(TP\), \(TN\), \(FP\), and \(FN\) denote true positives, true negatives, false positives, and false negatives, respectively.

\paragraph{Precision} 
Precision measures the fraction of correct POS tag predictions among all instances predicted as a given tag:
\begin{equation}
    \text{Precision} = \frac{TP}{TP + FP}.
\end{equation}

\paragraph{Recall} 
Recall determines the proportion of actual POS tag instances that were correctly predicted:
\begin{equation}
    \text{Recall} = \frac{TP}{TP + FN}.
\end{equation}

\paragraph{F1-score} 
The F1-score, representing the harmonic mean of precision and recall, is computed as:
\begin{equation}
    \text{F1-score} = 2 \times \frac{\text{Precision} \times \text{Recall}}{\text{Precision} + \text{Recall}}.
\end{equation}

\section{Further analysis of decoding strategies for POS tagging}
\label{a:further_analysis_pos_class}

Our analysis of gemma2\_9b's zero-shot POS tagging performance across three datasets (NAF, Chauliac, Lanfranco) reveals that Temperature 0.9 (temp\_09) achieves the best overall results, with highest average accuracy (0.7807) and weighted F1 score (0.7835).

\begin{table}[!ht]
\centering
\begin{tabular}{lccc}
\toprule
\textbf{Decoding Strategy} & \textbf{Avg. Accuracy} & \textbf{Avg. Weighted F1} & \textbf{Avg. Macro F1} \\
\midrule
temp\_09 & \textbf{0.7807} & \textbf{0.7835} & \textbf{0.5823} \\
temp\_03 & 0.7724 & 0.7736 & 0.5772 \\
p\_095 & 0.7686 & 0.7634 & 0.5690 \\
\bottomrule
\end{tabular}
\caption{Performance comparison of decoding strategies across all datasets}
\end{table}

Performance varies by dataset, with temp\_09 optimal for NAF and Chauliac, while temp\_03 performs best on Lanfranco. Notably, temp\_03 excels on more individual POS classes (21 vs. 14 for temp\_09), particularly with less frequent tags. Common POS tags (NOUN, ADP, CCONJ) consistently achieve high F1 scores (>0.8) across all strategies.

\paragraph{Class-specific performance} Our analysis reveals substantial performance disparities across POS classes. The model excels at tagging common categories like NOUN (F1>0.9 in Chauliac), ADP (F1>0.9 in Lanfranco), and CCONJ (F1>0.9 in Chauliac), while struggling significantly with INTJ (F1=0.0), PUNCT (F1 as low as 0.0017), PROPN (F1<0.1 in some settings), and AUX (F1<0.05 in Lanfranco). These challenging classes are typically underrepresented in the training data, highlighting a critical area for improvement. As expected, an avenue for enhancement would be to focus on gathering more examples of these challenging POS classes, which are often severely underrepresented in already low-resource settings. Targeted data augmentation or specialized few-shot learning approaches for these specific categories could potentially yield significant performance gains without requiring extensive additional resources. We recommend temp\_09 when prioritizing overall accuracy, while temp\_03 offers better performance for specific, particularly less frequent, POS classes.

\end{document}